\documentclass[cameraready]{Interspeech}
\ifcameraready
\title{Open ASR Leaderboard: Towards Reproducible and Transparent Multilingual and Long-Form Speech Recognition Evaluation}
\else
\title{ASR Leaderboard: Towards Reproducible and Transparent Multilingual and Long-Form Speech Recognition Evaluation}
\fi

\author[affiliation={1,5}, equalcontribution]{Vaibhav}{Srivastav}
\author[affiliation={1}, equalcontribution]{Steven}{Zheng}
\author[affiliation={1}, orcid=0000-0003-4837-5031]{Eric}{Bezzam}
\author[affiliation={1}]{Eustache}{Le Bihan}
\author[affiliation={2}, orcid=0009-0002-7371-4090]{Nithin Rao}{Koluguri}
\author[affiliation={2}, orcid=0000-0002-8245-0413]{Piotr}{Żelasko}
\author[affiliation={2}, orcid=0000-0001-5635-4893]{Somshubra}{Majumdar}
\author[affiliation={3}, orcid=0009-0007-3332-3526]{Adel}{Moumen}
\author[affiliation={4}]{Sanchit}{Gandhi}

\address{
    $^1$ Hugging Face, Inc., France\\
    $^2$ NVIDIA, United States of America\\
    $^3$ University of Cambridge, United Kingdom\\
    $^4$ Mistral AI, France\\
    $^5$ OpenAI, United States of America
}

\email{steven@huggingface.co, eric.bezzam@huggingface.co}

\keywords{benchmarking, automatic speech recognition, reproducible, multilingual, long-form}

\usepackage{comment}

\usepackage{amsmath,graphicx}
\usepackage[capitalize]{cleveref} % automatically formats references and capitalizes names like "Figure"
\usepackage{makecell}
\usepackage{booktabs}
\usepackage{subcaption}
\usepackage{caption}
\usepackage{hyperref}
\usepackage{xstring}
\usepackage{listings}
\usepackage{xfrac}
\usepackage{xcolor}

\newcommand{\eg}{\emph{e.g.},\ }   % for example
\newcommand{\ie}{\emph{i.e.},\ }   % that is
\definecolor{codebg}{RGB}{245, 245, 220} % beige
\newcommand{\rurl}[1]{%
  \StrBehind{#1}{://}[\display]%
  \edef\temp{\noexpand\href{#1}{\noexpand\texttt{\detokenize{\display}}}}%
  \temp
}

\begin{document}

\maketitle

\begin{abstract}
We present the \ifcameraready \textit{Open ASR Leaderboard}\else \textit{ASR Leaderboard}\fi, a reproducible benchmarking platform with community contributions from academia and industry. It compares 86 open-source and proprietary systems across 12 datasets, with English short- and long-form and multilingual short-form tracks. We standardize word error rate (WER) and inverse real-time factor (RTFx) evaluation for consistent accuracy-efficiency comparisons across model architectures and toolkits (\eg ESPNet, NeMo, SpeechBrain, Transformers). We observe that Conformer-based encoders paired with transformer-based decoders achieve the best average WER, while connectionist temporal classification (CTC) and token-and-duration transducer (TDT) decoders offer superior RTFx, making them better suited for long-form and batched processing. All code and dataset loaders are open-sourced to support transparent, extensible evaluation. We present our evaluation methodology to facilitate community-driven benchmarking in ASR and other tasks.
\end{abstract}

\section{Introduction}
\label{sec:intro}

Automatic speech recognition (ASR) has seen remarkable progress in recent years, fueled in part by open-source contributions.
Publicly-available datasets~\cite{librispeech,GigaSpeech2021} and pre-trained models~\cite{whisper,w2v2,hubert} have enabled researchers across academia and industry to build on existing work.
Yet, as the number of datasets and models grows, it becomes increasingly difficult for developers of new models to know which baselines to compare against and how. Similarly, users focused on inference may find it challenging to identify which model,
whether open-source or proprietary,
best meets their needs in terms of application and/or efficiency. 
Moreover, most existing benchmarks and evaluations overwhelmingly emphasize English and short-form transcription.

Several efforts have sought to address parts of this problem,
including benchmarks across multiple accents and diverse contexts in the French language~\cite{LeBenchmark},
under noise and reverberation in far-field settings~\cite{CHIME},
and comparing commercial and open-source models in English and Chinese~\cite{SpeechColab}.
Some common observations can be drawn from these efforts: (1)~there is no ``catch-all'' model, (2)~no single dataset is sufficient for evaluation, and (3)~a single metric, \ie word error rate (WER), is not enough.

To address these challenges,
we introduce the \ifcameraready \textit{Open ASR Leaderboard}\else \textit{ASR Leaderboard}\fi. Our contributions include:
\begin{enumerate}
\item An interactive leaderboard that compares $86$ open-source and proprietary models from $26$ organizations, with evaluations over $12$ datasets.\footnote{\ifcameraready\url{https://hf.co/spaces/hf-audio/open_asr_leaderboard}\else Link omitted for anonymity.\fi} Our comparison standardizes the evaluation across multiple open-source toolkits (ESPNet~\cite{espnet}, NeMo~\cite{nemo}, SpeechBrain~\cite{speechbrain1.0}, Transformers~\cite{transformers}), commercial APIs (AssemblyAI, Aqua Voice, Google, ElevenLabs, Rev AI, Speechmatics, Zoom), and model-specific repositories.
\item A multilingual benchmark covering German, French, Italian, Spanish, and Portuguese.
\item A dedicated evaluation for long-form transcription.
\end{enumerate}

For transparency and to facilitate the addition of new models and datasets, the evaluation scripts for the leaderboard are open-sourced.\footnote{\ifcameraready\url{https://github.com/huggingface/open_asr_leaderboard}\else Link omitted for anonymity.\fi}
The presented model, dataset, and languages count are as of 27 March 2026, and continue to increase with new additions to the leaderboard.

\ifcameraready
\section{Open ASR Leaderboard}
\else 
\section{ASR Leaderboard}
\fi
\label{sec:leaderboard}

\begin{table*}[t]
\caption{Datasets used for the \ifcameraready \textit{Open ASR Leaderboard}\else \textit{ASR Leaderboard}\fi. The \textit{Task(s)} column indicates which tasks (as denoted in \cref{sec:overview}) the dataset is used for. The duration is of the test-split. The \textit{Multilingual} datasets indicate the range of durations for the evaluated languages. Note that for \textit{Earnings22} a subset of \SI{5}{\hour} is used for short-form English (\textit{Leaderboard}) comparison.}
\label{tab:datasets}
\centering
\scalebox{0.7}{
\begin{tabular}{ccccccc}
\hline
\textbf{Dataset} & \textbf{Task(s)} & \textbf{Duration [h]} & \textbf{License} & \textbf{Source} & \textbf{Style} & \textbf{Transcriptions}\\
\hline
% https://huggingface.co/datasets/edinburghcstr/ami/viewer/ihm/test?views%5B%5D=ihm_test
% https://groups.inf.ed.ac.uk/ami/corpus/
AMI Meeting Corpus~\cite{ami} &  Leaderboard& 9 & CC-BY-4.0 & Meetings & Spontaneous & \makecell{Punctuated, cased, disfluencies} \\

CoVoST-2~\cite{covost} &  Multilingual  (de/fr/it/es/pt)& 
% FR - 23.3, DE - 21.5 , IT - 15.4, ES - 22.7 , PT - 5.326
5.3--23
& CC-BY-NC-4.0 & Open domain & Read & Punctuated, cased  \\

% https://lingtools.uoregon.edu/coraal/userguide/CORAALUserGuide_current.pdf
CORAAL~\cite{coraal} &  Long-form  & 159
& CC-BY-NC-4.0 & Sociolinguistic interviews & Spontaneous & \makecell{Punctuated, cased, disfluencies}  \\

% paper: https://arxiv.org/pdf/2104.11348
Earnings21~\cite{earnings21} &  \makecell{Long-form}&  39 & CC-BY-SA-4.0 & Earnings calls & Oratory, spontaneous & \makecell{Punctuated, cased, disfluencies} \\
% Earnings22 paper: https://arxiv.org/pdf/2203.15591
% linked on Leaderboard: https://huggingface.co/datasets/revdotcom/earnings22/blob/main/earnings22.py
% Another HF link: https://huggingface.co/datasets/distil-whisper/earnings22
Earnings22~\cite{earnings22} &  \makecell{Leaderboard, Long-form}& 119 & CC-BY-SA-4.0 & Earnings calls & Oratory, spontaneous & \makecell{Punctuated, cased, disfluencies}  \\
FLEURS~\cite{fleurs} &  Multilingual  (de/fr/it/es/pt)& 
% FR - 1.95, DE - 3.15, IT - 3.52, ES - 3.09, PT - 3.24
2.0--3.5 & CC-BY-4.0 & Wikipedia & Read & Punctuated, cased  \\
GigaSpeech~\cite{GigaSpeech2021} &  Leaderboard& 40 & apache-2.0 & Audiobook, podcast, YouTube & Read, spontaneous & \makecell{Punctuated, disfluencies}\\
% paper: https://www.danielpovey.com/files/2015_icassp_librispeech.pdf
LibriSpeech (clean)~\cite{librispeech} &  Leaderboard & 5.4 & CC-BY-4.0 & Audiobooks & Read & Normalized \\
LibriSpeech (other)~\cite{librispeech} &  Leaderboard&   5.1 & CC-BY-4.0 & Audiobooks (noisier) & Read &  Normalized  \\
% MLS page: https://www.openslr.org/94/
% FR - 5.94, DE - , IT - 2.68, ES - 6.33, PT - 0.79
MLS~\cite{mls} &  Multilingual (fr/it/es/pt) &   0.8--6.3 & CC-BY-4.0 & Audiobooks & Read &  Normalized  \\

SPGISpeech~\cite{spgispeech} &  Leaderboard&   100 & User Agreement & Financial meetings & Oratory, spontaneous & Punctuated, cased  \\

TED-LIUM v3~\cite{tedlium3} &  \makecell{Leaderboard, Long-form}&  3 & CC-BY-NC-ND 3.0 & TED Talks & Oratory & Disfluencies \\

VoxPopuli~\cite{voxpopuli} &  Leaderboard&  5 & CC0 & European Parliament & Oratory &Punctuated  \\
\hline
\end{tabular}}
\end{table*}

\subsection{Overview}
\label{sec:overview}

The \ifcameraready \textit{Open ASR Leaderboard} \else \textit{ASR Leaderboard} \fi contains evaluations on three tasks: 
\begin{enumerate}
    \item \textit{Leaderboard}, which evaluates short-form English transcription. We define short-form as audio less than \SI{30}{\second}, namely the receptive field of Whisper~\cite{whisper}.
    \item \textit{Multilingual}, which currently evaluates German, French, Italian, Spanish, and Portuguese transcription.
    \item \textit{Long-form}, which evaluates English transcription on audio longer than \SI{30}{\second}.
\end{enumerate}

A separate \textit{Long-form} evaluation is necessary because many recent models are derived from the pretrained encoder and/or architecture of Whisper (\SI{32}{\percent} of open models in our leaderboard; see \cref{tab:arch_breakdown}). Additionally, models may adopt different chunking strategies to reduce inference latency or employ different context window sizes during training (particularly for audio LLMs), both of which can impact long-form transcription quality.

The datasets used for evaluation are presented in~\cref{sec:datasets},
while evaluation metrics are described in \cref{sec:metrics}.
\cref{sec:current_models} provides an overview of the models evaluated within the \ifcameraready \textit{Open ASR Leaderboard}\else \textit{ASR Leaderboard}\fi,
while \cref{sec:adding_model} outlines the community-based process for contributing new models.

\subsection{Datasets}
\label{sec:datasets}

\cref{tab:datasets} summarizes the datasets used for the \ifcameraready \textit{Open ASR Leaderboard}\else \textit{ASR Leaderboard}\fi.
For the short-form evaluation (\textit{Leaderboard}), we segment the original audio into chunks of at most \SI{30}{\second}, with a small number of exceptions.
For normalized transcriptions, punctuation and casing is removed, as well as disfluencies such as fillers (\eg ``ah'', ``uh'', ``um''), repetitions, and repairs.
Some datasets retain a subset of these features in their provided transcriptions (as indicated in the last column of \cref{tab:datasets}).

While it is not possible to fully guarantee the absence of test-set contamination, 
namely that a given model has not been trained on a specific evaluation set \cite{evaluationllmsspeechflawed},
the use of multiple evaluation datasets per track enables us to identify anomalous performance on any single dataset relative to the others. 
In addition, each track includes at least one evaluation dataset released under a non-commercial license, which helps mitigate the risk of test-set contamination for commercial APIs and open-source models with commercial-friendly licenses: \textit{TED-LIUM v3} for \textit{Leaderboard}, \textit{CoVoST-2} for \textit{Multilingual}, and \textit{TED-LIUM v3} and \textit{CORAAL} for \textit{Long-form}.

Dataset retrieval and usage is enabled through the \textit{datasets} library~\cite{datasets}.
The datasets themselves are hosted on the Hugging Face Hub,\footnote{
\textit{Leaderboard}: \ifcameraready\url{https://hf.co/datasets/hf-audio/esb-datasets-test-only-sorted}\else Link omitted for anonymity\fi;\\\textit{Multilingual}: \ifcameraready\url{https://hf.co/datasets/nithinraok/asr-leaderboard-datasets}\else Link omitted for anonymity\fi;\\\textit{Long-form}: \ifcameraready\url{https://hf.co/datasets/hf-audio/asr-leaderboard-longform}; \url{https://huggingface.co/datasets/bezzam/coraal}\else Link omitted for anonymity.\fi}
which enables interactive exploration directly in the browser, including listening to individual audio, inspecting metadata, and running SQL queries, all without downloading the datasets.
The datasets can be conveniently downloaded and used in Python as such:

\begin{lstlisting}[language=Python, caption={Example of loading dataset.}]
from datasets import load_dataset

ds = load_dataset("hf-audio/esb-datasets-test-only-sorted", "ami", split="test")
audio_sample = ds[0]
\end{lstlisting}

\subsection{Metrics}
\label{sec:metrics}

We report results on two metrics: \textit{word error rate} (WER) for comparing transcription quality, and \textit{inverse real-time factor} (RTFx) for comparing inference speed.

Not all models produce transcripts with punctuation, casing, or disfluencies; in particular, some models explicitly remove the latter.
To account for discrepancies between model outputs and dataset transcriptions (last column of \cref{tab:datasets}),
we normalize all text prior to computing WER.
This normalization removes punctuation and casing, and applies an English text normalization pipeline closely following that of Whisper~\cite{whisper}.
The pipeline includes number normalization (\eg ``zero'' to ``0''), spelling standardization, and the removal of filler words.
On the leaderboard, models are sorted according to average WER across all datasets of a corresponding task.

We define RTFx as:
\begin{align}
    \text{RTFx} = \frac{\text{Total duration of audio}}{\text{Transcription time}}.
\end{align}
Higher values indicate faster inference (\ie lower latency).
We report the inverse real-time factor, rather than real-time factor (), so that ``higher is better'' and relative speedups are easy to interpret (\eg 10$\times$ or 100$\times$ faster).
RTFx can be computed for a single utterance or over a batch of audio.
RTFx (and related variants) is commonly used to quantify a model's efficiency on long-form audio~\cite{nvidia_longform}.
% As an aggregate score that accounts for transcription quality and efficiency, the ratio $\frac{\text{RTFx}}{\text{Avg. WER}}$ can be computed.
From the leaderboard page, models can be dynamically sorted by WER-performance on a particular dataset, or by RTFx.

\subsection{Current models}
\label{sec:current_models}

\begin{table}[t]
\centering
\caption{Distribution of encoder and decoder architectures for the open-source models in the \ifcameraready \textit{Open ASR Leaderboard}\else \textit{ASR Leaderboard}\fi. Some models use hybrid architectures for the encoder or decoder, and are counted twice. See \cref{sec:current_models} for the distinction between the encoder and decoder architectures.}
\scalebox{0.8}{
\begin{tabular}{lcccc|c}
\toprule
\textbf{Enc $\downarrow$ / Dec $\rightarrow$} & Transformer & CTC & RNN-T/TDT & LLM & Total \\
\midrule
Conformer-based   & 6  & 9  &  8 &  5 & 28 \\
Whisper     &  18 & 3  &  0 &  4 & 25 \\
Self-supervised   &  1 & 14  & 0  & 0  & 15 \\
Custom   & 7  &  0 &  0 &  3 & 10  \\
\midrule
Total             & 32 & 26 & 8 & 12 & 78 \\
\bottomrule
\end{tabular}
}
\label{tab:arch_breakdown}
\end{table}

%%% WITHOUT RANK COLUMN because annoying to update
\begin{table*}[t]
\caption{Subset of \ifcameraready \textit{Open ASR Leaderboard} \else \textit{ASR Leaderboard} \fi results on short-form English. WER is averaged over datasets corresponding to the \textit{Leaderboard} in \cref{tab:datasets}. \textit{Whisper-FT} stands for Whisper-finetuned. The top 10 are displayed along with additional models to comments on various architectures. The full and latest table can be found on Hugging Face.}
\centering
\scalebox{0.7}{
\begin{tabular}{lcccccc}
\toprule
\textbf{Model} & \textbf{Open} & \textbf{Avg. WER $\downarrow$} & \textbf{RTFx $\uparrow$} & \textbf{Encoder} & \textbf{Decoder} & \textbf{\# Lang.} \\
\midrule

Cohere Labs Transcribe & Yes & \textbf{5.42} & 525 & FastConformer~\cite{fastconformer} & Transformer & 14\\

Zoom Scribe v1 & No & 5.47 & - & - & - & 1\\

IBM Granite Speech 4.0 1B & Yes & 5.52 & 280 & Conformer~\cite{conformer} & LLM & 6\\

NVIDIA Canary Qwen 2.5B & Yes & 5.63 & 418 & FastConformer~\cite{fastconformer} & LLM & 1\\

IBM Granite Speech 3.3 8B & Yes & 5.76  &145 & Conformer~\cite{conformer} & LLM & 5\\

Qwen3 ASR 1.7B & Yes & 5.76  & 148 & Custom~\cite{Qwen3-ASR} & LLM & 52\\

ElevenLabs Scribe v2 & No & 5.83  & -- & -- & -- & 90+\\

IBM Granite Speech 3.3 2B & Yes & 6.00 &271 & Conformer~\cite{conformer} & LLM & 5\\

Microsoft Phi 4 Multimodal Instruct & Yes & 6.02 & 151 & Conformer~\cite{conformer} & LLM & 8\\

NVIDIA Parakeet TDT 0.6B v2 & Yes & 6.05 & 3390 & FastConformer~\cite{fastconformer} & TDT~\cite{tdt}  & 1\\

AssemblyAI Universal 3 Pro & No & 6.21 & -- & -- & --  & 99\\

% Aqua Voice Avalon & No & 6.24 & -- & -- & --  & 1\\

NVIDIA Parakeet TDT 0.6B v3 & Yes & 6.32 & 3330 & FastConformer~\cite{fastconformer} & TDT~\cite{tdt} & 25 \\

Google Chirp v2 & No & 6.42 & - & - & - & 468 \\

NVIDIA Canary 1B & Yes & 6.50 & 235 & FastConformer~\cite{fastconformer} & Transformer & 4 \\

Mistral AI Voxtral Small 24B & Yes & 6.62 & 54.1 & Whisper-FT~\cite{voxtral} & LLM & 8\\

Nyra Health CrisperWhisper & Yes & 6.67 & 84.1 & Whisper-FT~\cite{CrisperWhisper} & Whisper-FT & 1\\

Speechmatic Enhanced & No & 6.91 & -- & -- & -& 55 \\

RevAI Fusion & No & 7.12 & -- & -- & -- & 1\\

NVIDIA Canary 1B v2 & Yes & 7.15 & 749 & FastConformer~\cite{fastconformer} & Transformer & 25 \\

Distil-Whisper Large v3.5 & Yes & 7.21 & 202 & Whisper~\cite{whisper} & Transformer& 1 \\

NVIDIA Parakeet CTC 1.1B & Yes & 7.40 & 2730 & FastConformer~\cite{fastconformer} & CTC &1 \\

OpenAI Whisper Large v3 & Yes & 7.44 & 146 & Whisper~\cite{whisper} & Whisper&99 \\

OpenAI Whisper Large v3 Turbo & Yes & 7.83 & 200 & Whisper~\cite{whisper} & Whisper&99 \\

Meta Omnilingual ASR LLM 7B v2   & Yes & 8.14 & 66.0 & wav2vec2~\cite{w2v2} & Transformer & 1676 \\

NVIDIA FastConformer CTC Large & Yes & 8.96 & \textbf{6400} & FastConformer~\cite{fastconformer} & CTC&1 \\

\bottomrule
\end{tabular}}
\label{tab:results_overview}
\end{table*}

Of the 86 models currently listed in the \ifcameraready \textit{Open ASR Leaderboard } \else \textit{ASR Leaderboard }\fi (as of 27 March 2026), 74 are open-source.
The models come from 26 organizations: NVIDIA~(18), 
Meta/Facebook~(14), OpenAI~(8), Hugging Face~(5), Useful Sensors~(5), University of Washington~(4), ESPNet~(3), Google~(3), IBM~(3), Mistral AI~(3), Alibaba Cloud~(2), ElevenLabs~(2), Microsoft~(2), Rev AI~(2), SpeechBrain~(2), Applied Brain Research~(1), AssemblyAI~(1), Aqua Voice~(1), AssemblyAI~(1), Cohere Labs~(1), Kyutai~(1), Nyra Health~(1), Speechmatics~(1), Ultravox~(1), Z.ai~(1), and Zoom~(1).

A significant effort of this leaderboard has been to standardize the usage across several libraries: four commonly-used open-source toolkits (ESPNet, NeMo, SpeechBrain, Transformers), seven commercial APIs (AssemblyAI, Aqua Voice, Google, ElevenLabs, Rev AI, Speechmatics, Zoom), and model-specific repositories.
The evaluation scripts for each model are open-sourced.\footnote{\ifcameraready\url{https://github.com/huggingface/open_asr_leaderboard}\else Link omitted for anonymity.\fi}

\cref{tab:arch_breakdown} summarizes the encoder and decoder architectures used by the open-source models in the leaderboard.
The following encoder architectures are represented: Conformer-based encoders~\cite{fastconformer,conformer}, Whisper-based encoders~\cite{whisper},
self-supervised encoders (\ie wav2vec2~\cite{w2v2}, HuBERT~\cite{hubert}, data2vec~\cite{data2vec}),
and custom approaches~\cite{kyutai,moonshine,shi2026qwen3,liu2026voxtral,peng2026vibevoice}.
Whisper-based encoders either use the encoder model without modification~\cite{gandhi2023distilwhisper}, apply low-rank adaptation~\cite{lorawhisper}, fine-tune it~\cite{voxtral,CrisperWhisper}, or train from scratch~\cite{owsm_v4}. % GLM asr also trains from scratch I think? but no ref

The evaluated decoder architectures include transformer-based, CTC (Connectionist Temporal Classification), Recurrent Neural Network Transducer (RNN-T), Token-and-Duration Transducer (TDT)~\cite{tdt}, and LLM (Large Language Model)-based approaches.
Although both transformer-based and LLM-based decoders rely on the same underlying architecture and operate autoregressively, LLM-based decoders are pretrained on large-scale text corpora and can function as standalone text language models.

\subsection{Adding a new model}
\label{sec:adding_model}

External contributors can add a new model to the leaderboard by opening a \textit{pull request} (PR) that includes:
\begin{enumerate}
  \item A Python script for evaluation on a specific dataset, optionally specifying a model version.\footnote{Example scripts for each task can be found in: \ifcameraready\texttt{transformers/run\_eval.py}, \texttt{transformers/run\_eval\_ml.py}, \texttt{transformers/run\_whisper\_longform.py}\else links omitted for anonymity.\fi}
  \item A Bash script that calls the Python script for each dataset and model combination.\footnote{An example can be found in: \ifcameraready\\\texttt{transformers/run\_whisper.sh}\else link omitted for anonymity.\fi}
  \item Self-reported metrics.
\end{enumerate}

We verify these scripts in our environment to ensure consistent evaluation across models before updating the leaderboard.
To date, 29 PRs have been merged to add 56 models following this process, and an additional 34 PRs have been merged to address other aspects and fixes.
% https://docs.google.com/spreadsheets/d/1woIjD5wwew1VTp1qd_MAXScfMKUsGDpvJPcSaOhsgIs/edit?gid=1087682557#gid=1087682557

\section{Results}
\label{sec:results}

\begin{figure*}[t]
  \centering
  \begin{subfigure}[t]{0.48\textwidth}
    \centering
    \includegraphics[width=\linewidth]{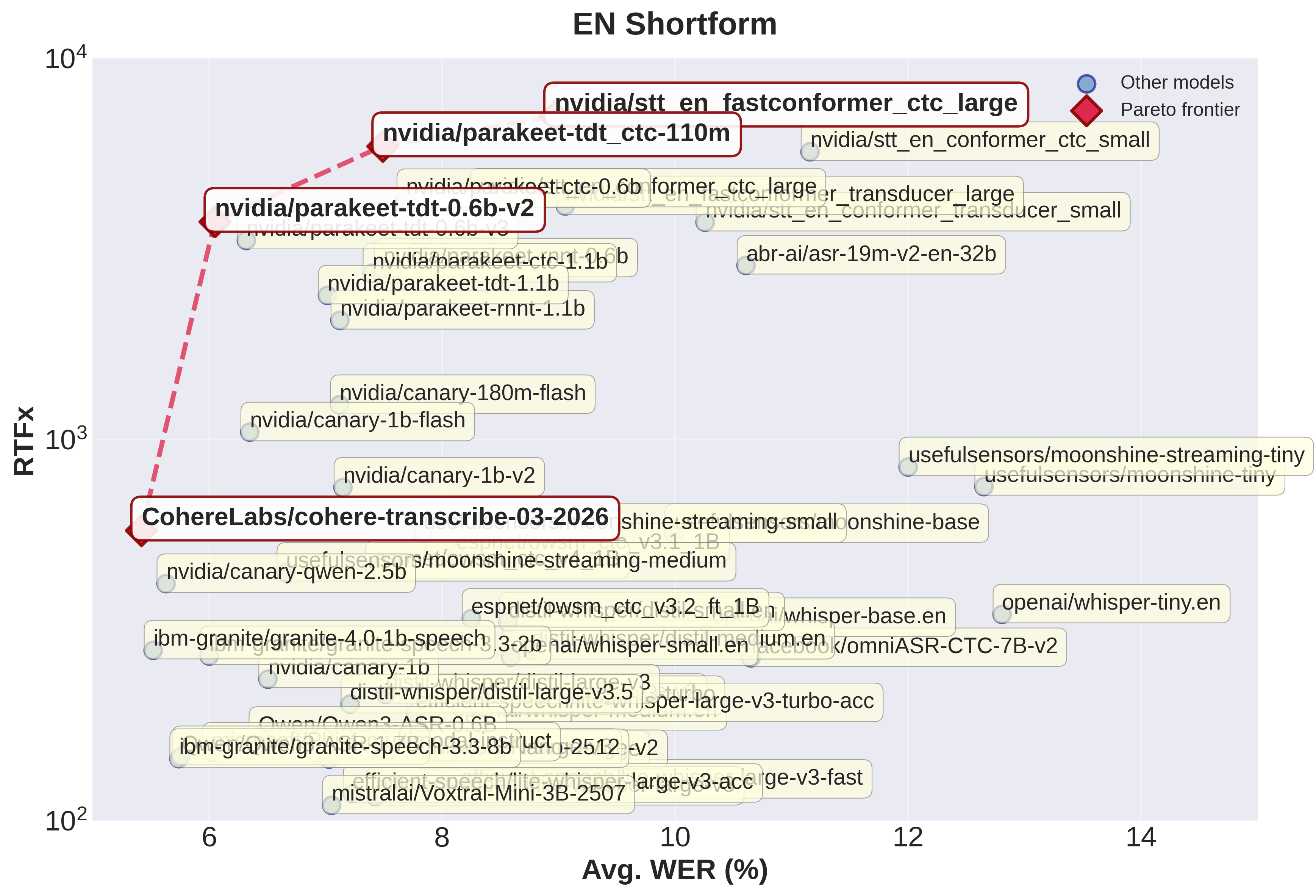}
    \caption{Average WER (\%) vs.\ RTFx}
    \label{fig:rtfx_wer}
  \end{subfigure}
  \hfill
  \begin{subfigure}[t]{0.48\textwidth}
    \centering
    \includegraphics[width=\linewidth]{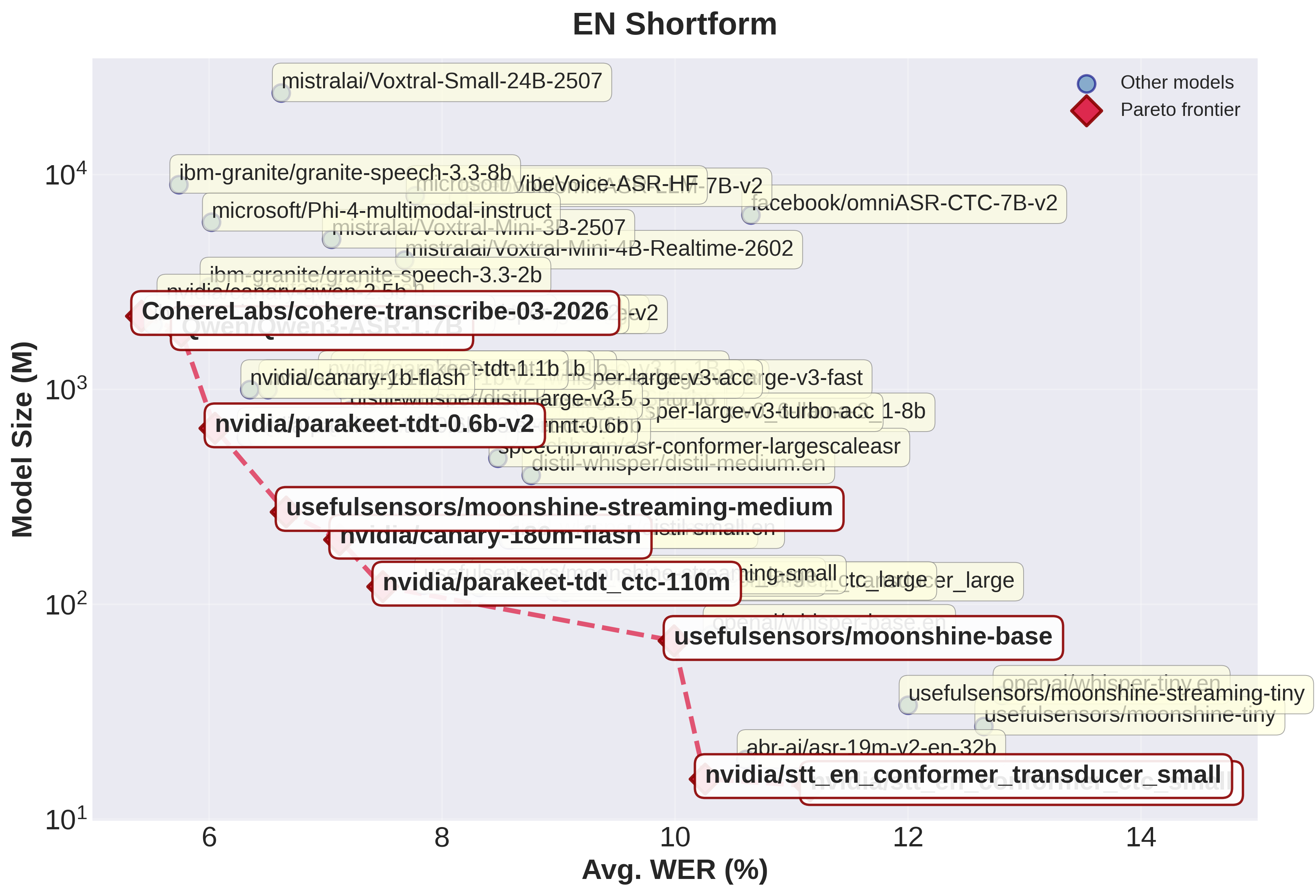}
    \caption{Average WER (\%) vs.\ Model Size (M)}
    \label{fig:size_wer}
  \end{subfigure}
  \caption{(a) Average word error rate (WER) vs.~inverse real time factor (RTFx) and (b) WER vs.~model size trade-off plots for the open-source ASR models, on short-form English transcription. See \cref{tab:results_overview} for metrics.}
  \label{fig:wer_tradeoff}
\end{figure*}

The evaluation scripts for each model were run on an NVIDIA A100-SXM4-80GB GPU (driver 560.28.03, CUDA 12.6), using a batch size of 64 whenever memory allowed, and reduced adaptively (48, 32, 16, \dots) when necessary to fit in device memory.
Although the absolute RTFx values depend on the underlying hardware and can vary substantially across systems, all measurements reported here are obtained under the same setup. As such, they provide a meaningful basis for comparing the relative efficiency of models, even if the absolute numbers may not directly transfer to other hardware configurations.

Since the full results are continuously updated on the \ifcameraready \textit{Open ASR Leaderboard} \else \textit{ASR Leaderboard} \fi and are too extensive to fully include here, we present a condensed versions of the English leaderboard in \cref{tab:results_overview}, the multilingual results in \cref{tab:wer_languages}, and the long-form results in \cref{tab:longform_results}. The full leaderabords can be found on Hugging Face: \url{https://hf.co/spaces/hf-audio/open_asr_leaderboard}

\subsection{Short-form (English)}

Models that combine a Conformer-based encoder~\cite{fastconformer,conformer} with a transformer-based decoder (either custom or LLM-based) achieve the best average WER. However, these models are significantly slower than those using TDT or CTC decoders.
While TDT/CTC approaches offer substantially better RTFx, this speed advantage comes at the expense of accuracy: the highest-ranking TDT model (\textit{NVIDIA Parakeet TDT 0.6B v2}) places 10th, while the best CTC model (\textit{NVIDIA Parakeet CTC 1.1B}) ranks 34th.

As shown in \cref{tab:arch_breakdown}, Conformer-based encoders are the most widely adopted. Restricting the comparison to models with transformer/LLM-based decoders (10 Conformer-based and 22 Whisper-based), Conformer-based systems are on average $3.77\times$ faster, with an average RTFx of 758 compared to 201 for Whisper-based models.

Despite this, Whisper-based encoders remain popular due to pretraining on large-scale multilingual data, enabling support for up to 99 languages. Models that fine-tune Whisper's encoder for specific languages (\eg \textit{Nyra Health CrisperWhisper}~\cite{CrisperWhisper} and \textit{Mistral AI Voxtral Small 24B}~\cite{voxtral}), or that train a new decoder (\textit{Distil-Whisper Large v3.5}~\cite{gandhi2023distilwhisper}) can achieve better average WER than \textit{OpenAI Whisper Large v3}.
Self-supervised encoders make up \SI{19}{\percent} of the approaches in the benchmark. While they enable systems for 1600+ languages~\cite{omnilingual2025omnilingual}, the best approach ranks only 53rd (\textit{Meta Omnilingual ASR LLM 7B v2}).

The scatter plots in \cref{fig:wer_tradeoff} show all open-source models with a WER below \SI{15}{\percent} plotted against RTFx and model size.
The Pareto front is overlaid to illustrate the trade-offs between WER and each of these criteria.

\subsection{Multilingual}

\begin{table}[t]
\caption{Average WERs for each language (German/French/Italian/Spanish/Portuguese) and across all languages on the \textit{Multilingual} datasets in \cref{tab:datasets}. The full and latest table on Hugging Face.}
\centering
\scalebox{0.6}{
\begin{tabular}{lcccccccc}
\toprule
\textbf{Model} & \textbf{Open} & \makecell{\textbf{Avg.}\\\textbf{WER}} & \textbf{RTFx} & \textbf{DE} & \textbf{FR} & \textbf{IT} & \textbf{ES} & \textbf{PT} \\
\midrule

ElevenLabs Scribe v2  & No & \textbf{2.67} & -- & \textbf{2.27} & \textbf{3.28} & \textbf{2.58} & \textbf{2.33} & \textbf{2.83} \\

Assembly AI Universal 3 Pro  & No & 3.23 & -- & 2.34 & 3.74 & 3.94 & 2.34 & 3.63 \\

Mistral AI Voxtral Small 24B  & Yes & 3.70 & 42.0 & 3.01 & 4.13 & 3.91 & 3.04 & 4.40\\

Cohere Labs Transcribe  & Yes & 3.83 & 491 & 3.84 & 4.05 & 3.44 & 2.81 & 5.60 \\

Speechmatic Enhanced   & No & 4.29 & -- & 2.84 & 5.04 & 5.58 & 2.78 & 4.97 \\

Meta Omnilingual ASR LLM 7B v2  & Yes & 4.39 & 21.2 & 4.55 & 5.34 & 3.75 & 3.44 & 5.18 \\

Microsoft Phi 4 Multimodal Instruct   & Yes & 4.41 & 78.2 & 3.96 & 5.20 & 4.15 & 3.71 & 5.12 \\

NVIDIA Canary 1B v2  & Yes & 4.60 & 634 & 4.10 & 4.83 & 4.88 & 3.25 & 6.33 \\

Meta Omnilingual ASR LLM 7B  & Yes & 4.68 & 21.8 & 4.58 & 5.46 & 3.80 & 3.73 & 6.36 \\

OpenAI Whisper Large v3  & Yes & 4.81 & 111 & 4.26 & 6.36 & 4.69 & 3.65 & 4.96 \\

NVIDIA Parakeet TDT 0.6B v3   & Yes & 4.81 & \textbf{1720} & 4.20 & 5.42 & 4.81 & 3.73 & 6.16\\

Qwen3 ASR 1.7B  & Yes & 5.11 & 113 & 4.12 & 5.74 & 5.61 & 3.87 & 6.29\\

Meta Omnilingual ASR CTC 7B v2  & Yes & 5.84 & 155 & 6.05 & 7.63 & 4.79 & 4.49 &6.57 \\

% Voxtral Mini 4B Realtime  & Yes & 6.32 & 28.2 & 6.56 & 7.92 & 6.63 & 4.27 & 6.29 \\

\bottomrule
\end{tabular}}
\label{tab:wer_languages}
\end{table}

\begin{figure*}[t]
  \centering
  \begin{subfigure}[t]{0.48\textwidth}
    \centering
    \includegraphics[width=\linewidth]{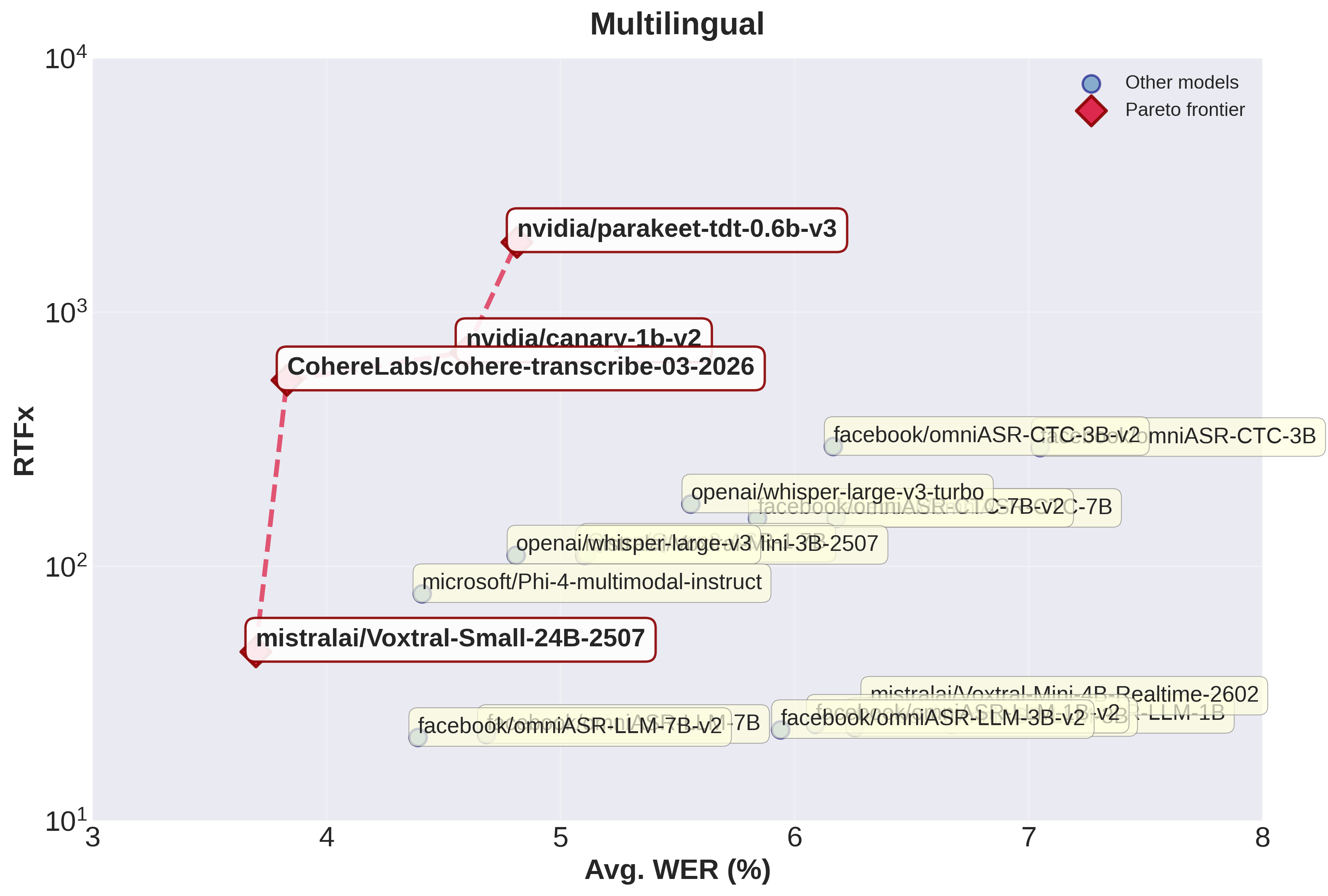}
    \caption{Multilingual: Average WER (\%) vs.\ RTFx}
    \label{fig:multilingual_rtfx_wer}
  \end{subfigure}
  \hfill
  \begin{subfigure}[t]{0.48\textwidth}
    \centering
    \includegraphics[width=\linewidth]{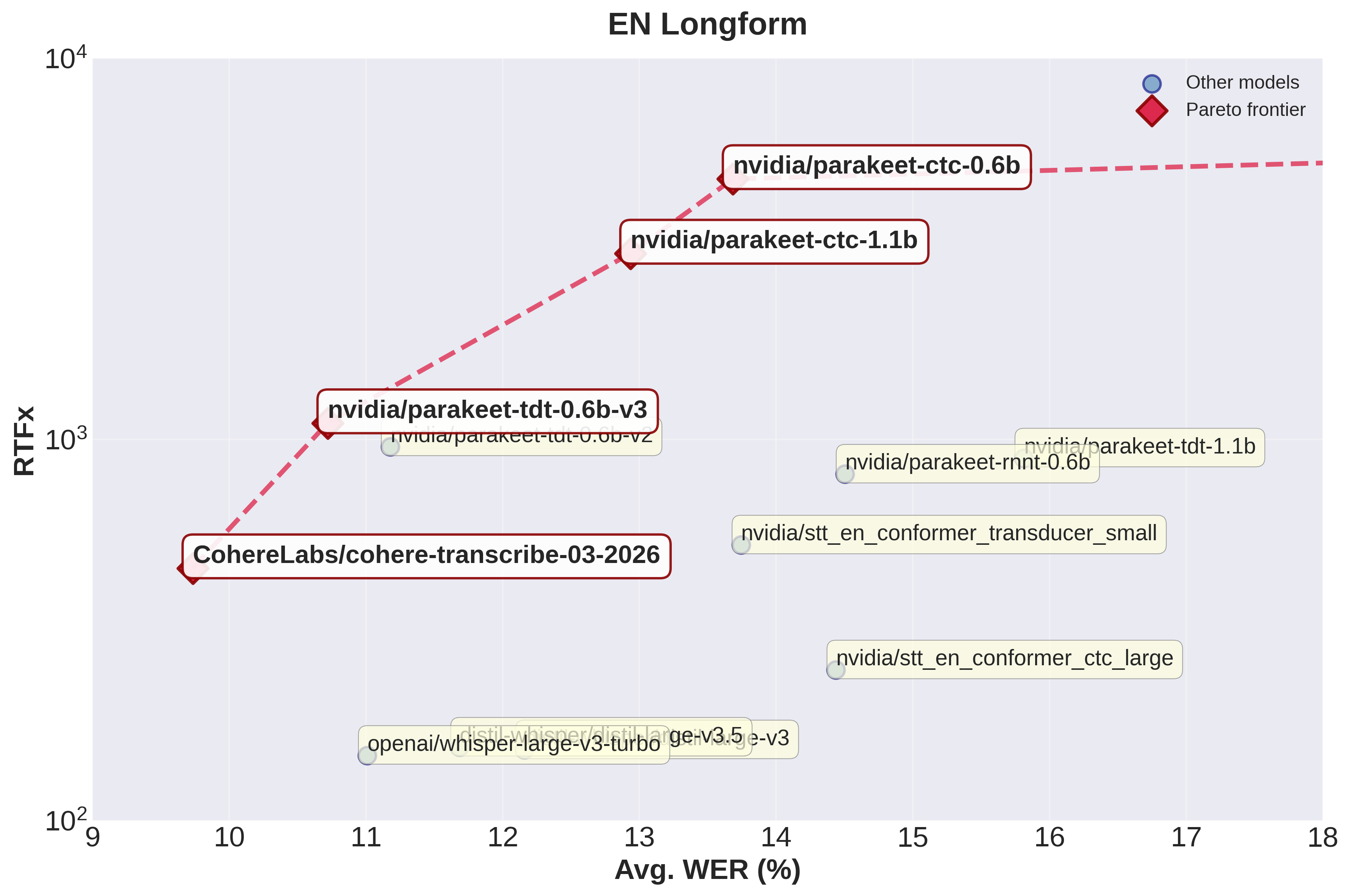}
    \caption{Average WER (\%) vs.\ Model Size (M)}
    \label{fig:longform_rtfx_wer}
  \end{subfigure}
  \caption{Average word error rate (WER) vs.~inverse real time factor (RTFx) for(a) multilingual and (b) longform. See \cref{tab:wer_languages} and \cref{tab:longform_results} respectively for metrics.}
  \label{fig:multilingual_longform}
\end{figure*}

Closed-source models achieve the best results on the multilingual datasets (see \cref{tab:wer_languages}).
Among open-source models, we observe a trade-off between specialization and broad multilingual coverage. 
With a large and diverse set of models, NVIDIA's systems provide a clear example of this trade-off: \textit{Parakeet TDT 0.6B v3} adds multilingual support compared to v2, and \textit{Canary 1B v2} expands from 4 to 25 languages. In both cases, broader language coverage comes at the cost of English transcription accuracy. Similarly, \textit{Meta Omnilingual ASR LLM 7B v2} ranks higher on the multilingual benchmarks, outperforming several models that ranked higher on the short-form English benchmark.
\cref{fig:multilingual_rtfx_wer} plots all open-source models with a WER below \SI{8}{\percent} against RTFx.
% Namely, the improvement in English performance of fine-tuned Whisper models often comes at the cost of multilingual coverage: Whisper-derived models typically train on fewer languages (often just English), while Whisper itself supports 99 languages.

\subsection{Long-form (English)}

\begin{table}[t]
\caption{Subset of results on the \textit{Long-form} datasets of \cref{tab:datasets}. The full and latest table is on Hugging Face.} 
\centering
\scalebox{0.7}{
\begin{tabular}{lccc}
\toprule
\textbf{Model} & \textbf{Open} & \textbf{Avg. WER} & \textbf{RTFx}  \\
\midrule
ElevenLabs Scribe v2 & No & 7.32 & -- \\
AssemblyAI Universal 3 Pro & No & 8.34 & -- \\
Speechmatics Enhanced & No & 8.80 & --  \\
RevAI Fusion & No & 9.54 & -- \\
% RevAI Machine & No & 9.64 & -- \\
Cohere Labs Transcribe & Yes & 9.73 & 418  \\
NVIDIA Parakeet TDT 0.6B v3 & Yes & 10.7 & 1000  \\
OpenAI Whisper Large v3 Turbo & Yes & 11.0 & 148 \\
% NVIDIA Parakeet TDT 0.6B v2 & Yes & 11.2 & 956  \\
NVIDIA Canary Qwen 2.5B & Yes & 11.2 &  16.1 \\
OpenAI Whisper Large v3 & Yes & 11.2 & 68.6 \\
Distil-Whisper Large v3.5 & Yes & 11.7 & 156 \\

NVIDIA Parakeet CTC 1.1B & Yes & 12.9 & 2790 \\
Google Chirp & No & 13.0 & -- \\
NVIDIA Parakeet CTC 0.6B & Yes & 13.7 & \textbf{4383} \\
% NVIDIA FastConformer CTC Large & Yes & 21.5 & \textbf{5530} \\
\bottomrule
\end{tabular}}
\label{tab:longform_results}
\end{table}

Closed-source models also deliver the strongest results on long-form English, with a distinct gap over open-source alternatives (\cref{tab:longform_results}). Although the exact reasons are not known, this may be due to domain-specific fine-tuning. Moreover, differences between short-form and long-form performance may also arise from factors such as model context size, audio chunking strategies, and the handling of disfluencies, which tend to occur more frequently in long-form recordings (\eg meetings, presentations, and interviews).
% Among open-source models, \textit{NVIDIA Parakeet TDT 0.6B v3} performs best. Consistent with the short-form setting, CTC- and TDT-based architectures offer significantly higher throughput, making them particularly well suited for large-scale or batch transcription.
\cref{fig:longform_rtfx_wer} plots all open-source models with a WER below \SI{18}{\percent} against RTFx.

\section{Conclusions}
\label{sec:conclusion}

We present the \ifcameraready \textit{Open ASR Leaderboard}\else \textit{ASR Leaderboard}\fi, a reproducible benchmark covering $86$ systems and $12$ datasets, including multilingual and long-form speech. 
Our comparison standardizes the evaluation across multiple open-source toolkits (ESPNet, NeMo, SpeechBrain, Transformers), commercial APIs, and model-specific repositories.
Standardized text normalization enables a unified basis for comparing WER performance accuracy, and our RTFx evaluation allows for efficiency comparisons.
Conformer-based encoders paired with transformer-based decoders achieve the strongest English WER but at the cost of higher latency. In contrast, CTC- and TDT-based decoders offer faster inference with only modest accuracy trade-offs, making them attractive for long-form transcription. Code and datasets are open-sourced to support transparent and extensible evaluation.

Future work includes expanding evaluations across languages and domains (\eg far-field speech), incorporating additional metrics (\eg token error rate~\cite{SpeechColab}), and exploring underrepresented encoder-decoder combinations (\cref{tab:arch_breakdown}). With the rise of LLMs and their demonstrated strength in ASR, we anticipate more approaches leveraging them. To further improve benchmark reliability, private evaluation sets could help minimize the risk of test-set contamination.
Additionally, because models vary in how they handle disfluencies, it may be valuable to differentiate tasks based on whether disfluencies are explicitly modeled or whether verbatim transcription is required.

\ifcameraready
\section{Acknowledgments}
The authors would like to thank all the contributors to the \textit{Open ASR Leaderboard}.
\fi

\bibliographystyle{IEEEtran}
\bibliography{mybib}

\end{document}